\documentclass[]{spie}  

\usepackage{amsmath,amsfonts,amssymb}
\usepackage{graphicx}
\usepackage[colorlinks=true, allcolors=blue]{hyperref}
\usepackage{bm}
\usepackage{booktabs} 
\usepackage{multirow} 
\usepackage{float}

\title{Structure-Constrained Optimal Transport for Pulmonary Vessel Point Cloud Registration in Longitudinal Chest CT}

\author[a, *]{Chou Mo}
\author[b, *]{Yunzheng Zhu}
\author[b]{Luoting Zhuang}
\author[b]{William Hsu}
\affil[a]{Department of Mathematics, University of California, Los Angeles, Los Angeles, USA}
\affil[b]{Medical \& Imaging Informatics, Department of Radiological Sciences, David Geffen School of Medicine at University of California, Los Angeles, Los Angeles, USA}

\authorinfo{*Equal Contribution}

\begin{document} 
\maketitle

\begin{abstract}
Deformable registration of longitudinal chest CT is essential for tracking pulmonary nodules and assessing disease progression, where the pulmonary vessels provide a dense, physiologically meaningful set of correspondences for recovering respiratory motion. Optimal transport (OT) frames deformable registration as a distribution-matching problem, establishing correspondence between two point sets through a transport plan, and thus offers an alternative to intensity- and learning-based registration. We adopt a training-free Iterative OT formulation that aligns vessel centerline point clouds through a smooth polynomial warp, alternating between an OT correspondence step and a warp-fitting step. However, when different vascular structures, such as pulmonary arteries and veins, are registered as a single combined point cloud, the OT plan suffers from anatomically invalid cross-structure correspondence that degrades the fitted deformation. To address this, we propose a structure-constrained correspondence approach. At each iteration, we solve a separate OT within each anatomical structure, while fitting a single shared deformation jointly to all structures. On 308 intra-patient pairs from 207 National Lung Screening Trial subjects, the method significantly reduces nodule target registration error (TRE) at both point budgets ($p<0.001$). At a budget of 1000 points, it lowers nodule TRE from 2.73 to 2.49 mm ($8.8\%$), raises lung Dice from 0.9516 to 0.9537 and vessel Dice from 0.7426 to 0.7497, and converges in fewer iterations (42.2 to 27.9) while reducing fit time from 1.46 to 0.93 seconds. By restricting the transport to anatomically valid, smaller subproblems, the method is more accurate and faster than the combined baseline at equal point budget. 
\end{abstract}


\section{INTRODUCTION}
\label{sec:intro} 
Longitudinal image registration establishes a spatial correspondence between scans of the same subject acquired at different times, enabling quantitative comparison of anatomy across the interval. In thoracic imaging, longitudinal registration of chest computed tomography (CT) underlies pulmonary nodule tracking and the assessment of parenchymal disease, where anatomy must be matched across scans to measure change \cite{Cunliffe2012Lung,Shi2007Pulmonary}. Lung registration is challenging because the lung undergoes large nonrigid deformation between acquisitions, driven primarily by differences in respiratory state. Therefore, the transformation to be estimated is high-dimensional and spatially heterogeneous. Traditional intensity-based methods estimate a dense deformation field by optimizing an image similarity metric, but their accuracy degrades under large deformation and in regions of low intensity contrast such as the lung interior \cite{Stavropoulou2021A}. Learning-based methods regress the deformation directly but require large training sets of paired or annotated images, which are costly to assemble for longitudinal cohorts \cite{Balakrishnan2019VoxelMorph,Fu2020Deep}. Optimal transport (OT), a growing branch of low-cost point cloud registration \cite{Besl1992Method, Myronenko2010Point, Hansen2019Learning}, forms correspondence through sparse point set distribution matching. Instead of estimating the transformation from a single transport plan, iterative optimal transport (IOT)\cite{Li2026Iterative} alternates between a regularized unbalanced OT correspondence and a polynomial transformation, which inherently refines the transport plan and the transformation with block-coordinate descent under a biconvex objective. However, IOT was only formulated for two-dimensional multimodal registration and does not encode the anatomical information of the structures being matched. To address these issues, we 1) extend the problem to three-dimensional (3D) vascular structures of chest CT, and 2) develop structure-constrained correspondence for distinct structures, such as pulmonary arteries and veins.

Our proposed structure-constrained correspondence IOT contributes as follows: 1) We reduced the possibility of incorrect matches from distinct anatomical structures (pulmonary arteries and veins) and the iterations for convergence, while improving the registration accuracy, such as nodule TRE (2.73 to 2.49 mm), Dice scores of lung (0.9516 to 0.9537) and vessel (0.7426 to 0.7497), respectively; 2) we implemented a GPU-supported framework, which largely reduces the computational cost for processing the 3D chest CT vascular structures.

\section{METHOD}
\label{sec:method}
\subsection{Data}
\label{sec:method:data}
We evaluate on intra-patient longitudinal chest CT pairs from the National Lung Screening Trial (NLST). The dataset includes 308 pairs from 207 patients by permuting two of the three scans from each patient. Each pair comprises two screening acquisitions of the same subject. For every scan, a pulmonary nodule centroid (generated by nodule detection algorithms \cite{cardoso2022monai,liao2019evaluate} and confirmed with human annotators) is provided as a world-coordinate landmark and used only for evaluation. 

Lung, pulmonary arteries and veins are segmented with TotalSegmentator 2.13\cite{Jakob2023TotalSegmentator,Liu2025A,Skeleton2024}, yielding a separate binary mask per structure. Masks of pulmonary arteries and veins are thinned to their centerlines, allowing each vessel to be represented by a one-voxel-wide curve along its center. Arteries and veins are skeletonized separately as distinct trees, and bridged with joint thinning. The centerline voxels are converted to physical millimeter coordinates and restricted to the lung mask, giving one point cloud per structure. For the combined-cloud baseline, the two structures' centerlines are unioned as one point cloud. 

\subsection{Iterative Optimal-Transport Registration}
\label{sec:method:iot}

Similar to the framework of IOT \cite{Li2026Iterative}, a fixed point cloud $\mathbf{X}=\{\mathbf{x}_i\}_{i=1}^{N_1}\in\mathbb{R}^{N_1\times d}$ is registered to a moving point cloud $\mathbf{Y}=\{\mathbf{y}_j\}_{j=1}^{N_2}\in\mathbb{R}^{N_2\times d}$ ($d=3$) by a polynomial transformation $\mathcal{T}_{\bm{\theta}}$. Each point is lifted into a degree-$p$ polynomial feature map $\bm{\phi}(\mathbf{x})\in\mathbb{R}^{P}$ and transported as
\begin{equation}
  \mathcal{T}_{\bm{\theta}}(\mathbf{x}) = \bm{\phi}(\mathbf{x})^{\!\top}\bm{\theta},
  \qquad \bm{\theta}\in\mathbb{R}^{P\times d},
  \label{eq:warp}
\end{equation}
where $\bm{\phi}$ collects the monomials $\{x_a\}$, $\{x_ax_b\}_{a<b}$,
$\{x_a^q\}_{q=2}^{p}$ and a constant term, where $P$ is the number of such terms. For $d=3,\,p=3$, $\mathbf{x}=(x,y,z)$ and $\bm{\phi}(\mathbf{x}) = [\,x,\ y,\ z,\ xy,\ xz,\ yz,\ x^2,\ y^2,\ z^2,\ x^3,\ y^3,\ z^3,\ 1\,]^{\!\top}$, so $P=13$ and $\bm{\theta}$ is a single $13\times 3$ deformation of all of $\mathbb{R}^3$. Both clouds are standardized by the fixed-cloud per-axis mean and standard deviation before fitting. Registration alternates two steps by block-coordinate descent: a \emph{correspondence step} that updates a soft transport plan $\mathbf{W}$, a soft matching between the two clouds, and a \emph{warp step} that updates the transformation parameters $\bm{\theta}$. The two steps are iterated until the plan and the warp stop changing between consecutive iterations, $\max(\lVert\Delta\mathbf{W}\rVert_F,\lVert\Delta\bm{\theta}\rVert_F)<10^{-6}$, or a fixed iteration cap is reached.

\subsection{Structure-Constrained Correspondence}
\label{sec:method:corr}

Let each point carry an anatomical structure label $s\in\{1, \dots, S\}$ (for the scope of this paper, $S=2$ for artery and vein), partitioning the clouds into $\mathbf{X}=\bigcup_s \mathbf{X}_s$ and $\mathbf{Y}=\bigcup_s \mathbf{Y}_s$. Instead of building a single plan over the pooled cloud, we compute one unbalanced OT (UOT) correspondence within each structure,
\begin{equation}
  \mathbf{W}_s = \operatorname{UOT}\!\bigl(M^{(s)}_{ij}\bigr),\qquad
  M^{(s)}_{ij}=\lVert \mathcal{T}_{\bm{\theta}}(\mathbf{x}^{(s)}_i)-\mathbf{y}^{(s)}_j\rVert_2^2,
  \qquad s=1,\dots,S,
  \label{eq:blockdiag}
\end{equation}
where $\mathbf{X}_s$ and $\mathbf{Y}_s$ are the fixed and moving points of structure $s$. Solving these $S$ independent problems is equivalent to constraining a single transport plan over the pooled cloud to be \emph{block-diagonal}, with the off-diagonal (cross-structure) blocks fixed to zero. We use this block-diagonal form only as the conceptual link to the pooled baseline. In practice, we simply keep the $S$ per-structure plans $\mathbf{W}_s$ (each of size $N_s\times M_s$) and never form the $N_1\times N_2$ matrix.

\subsection{Shared Deformation via Joint Warp Step}
\label{sec:method:shared}
A single transformation $\bm{\theta}$ is fit to all structures' correspondences at every iteration, using the same closed-form step on stacked inputs. With $\bm{\Phi}_s=\bm{\phi}(\mathbf{X}_s)$ and the vertical concatenations $\bm{\Phi}_{\mathrm{all}}=[\bm{\Phi}_1;\dots;\bm{\Phi}_S]$,
$\mathbf{X}_{\mathrm{all}}=[\mathbf{X}_1;\dots;\mathbf{X}_S]$,
$\mathbf{w}_{\mathrm{all}}=[\mathbf{W}_1\mathbf{1};\dots;\mathbf{W}_S\mathbf{1}]$, and
$\mathbf{W}\mathbf{Y}=[\mathbf{W}_1\mathbf{Y}_1;\dots;\mathbf{W}_S\mathbf{Y}_S]$,
\begin{equation}
  \bm{\theta} =
  \bigl(\bm{\Phi}_{\mathrm{all}}^{\!\top}\!\operatorname{diag}(\mathbf{w}_{\mathrm{all}})\,
        \bm{\Phi}_{\mathrm{all}}
      + g\,\bm{\Phi}_{\mathrm{all}}^{\!\top}\bm{\Phi}_{\mathrm{all}}\bigr)^{-1}
  \bm{\Phi}_{\mathrm{all}}^{\!\top}\bigl(\mathbf{W}\mathbf{Y} + g\,\mathbf{X}_{\mathrm{all}}\bigr),
  \qquad g=\frac{\lambda}{N_{\mathrm{tot}}\,d},
  \label{eq:tstep-multi}
\end{equation}
with $N_{\mathrm{tot}}=\sum_s N_s$. Each $\mathbf{W}_s$ multiplies only its own $\mathbf{Y}_s$, so that the correspondence remains block-diagonal, while $\mathbf{A}$, $\mathbf{B}$ and $\bm{\theta}$ mix all structure, and one deformation is fit to every structure jointly.

For the joint fit to be well posed, all structures are standardized by the same statistics (the per-axis mean and standard deviation of the union of the fixed points), allowing the single polynomial map to be expressed in one common coordinate frame. Furthermore, the structures remain coupled throughout the optimization. After each warp step~\eqref{eq:tstep-multi}, every structure is warped again by updated transformation, $\mathbf{v}^{(s)}_i=\mathcal{T}_{\bm{\theta}}(\mathbf{x}^{(s)}_i)$, before its next correspondence~\eqref{eq:blockdiag} is computed. This ensures that each structure's subsequent matching is evaluated on points that all structures jointly displaced. Therefore, the structures constrain one another continuously through the shared deformation, and the correspondence is resolved per structure.

%

\section{EXPERIMENTS}
\label{sec:exp}
\subsection{Experiment Setup and Evaluation Metrics}
\label{sec:exp:expsetup}
We sampled $2000$ points with Farthest Point Sampling (FPS) from the centerlines of each structure. Then, we applied a global rigid transformation $\mathcal{T}_0$ per pair to ensure that the iterative registration operates on non-rigid motion rather than table-position offset. 
For a total budget $N$, the structure-constrained method draws $N/S$ points per structure by FPS, and the combined-cloud baseline draws $N$ points from the union point clouds. The unbalanced OT relaxation $\rho=2\times 10^{-2}$, and the identity regularizer weight $\lambda=10^{-3}$. The loop runs to convergence with a cap of 100 iterations and inner unbalanced OT tolerance $10^{-6}$. All fits use a GPU implementation that builds on the reference solver of Li et al. \cite{Li2026Iterative} and departs from it in two respects: (1) The warp step is solved in closed form~\eqref{eq:tstep-multi} instead of the original quasi-Newton inner loop, which is exact because the cost is quadratic in $\bm{\theta}$ and the two therefore share the same minimizer. (2) The cost matrix and the unbalanced OT updates run on a single NVIDIA Tesla V100-SXM2 GPU (32 GB). 

The baseline is the combined-cloud correspondence, where arteries and veins are pooled into one cloud and matched by a single unbalanced OT plan, with all other computations (pre-alignment, point budget, warp degree, relaxation, and stopping rule) held constant to the structure-constrained method (exactly the $S=1$ instance). 
\label{sec:exp:metrics}
We report three accuracy metrics and two efficiency metrics, computed identically for both methods. Nodule target registration error (TRE) is the Euclidean distance between the warped fixed nodule landmark and the moving landmark.  Lung Dice and vessel Dice (the union of artery and vein mask) are computed on a common $1$~mm-isotropic world lattice over a margin-dilated bounding box, following the GLIDE-Reg protocol\cite{glidereg}. The lattice is uniformly sub-sampled for an unbiased Dice estimate. For efficiency, we report the fit-only wall clock time and the number of iterations to convergence.

\section{RESULTS}
\begin{table}[t]
  \centering
  \caption{Registration accuracy, efficiency, and paired comparison on the NLST nodule cohort (308 pairs, 207 patients), averaged over pairs, at matched total point budget $N$. The first two rows are before-registration references. Arrows indicate the improving direction, and the better value per metric is in bold. Win and $p$ (paired $t$-test on nodule TRE) compare the structure-constrained method against the combined baseline at the same $N$.}
    \label{tab:results}
    \setlength{\tabcolsep}{4pt}
    {\small
    \begin{tabular}{llccccccc}
      \toprule
    $N$ & Method & TRE (mm)\,$\downarrow$ & Lung Dice\,$\uparrow$ & Vessel Dice\,$\uparrow$ & Fit (s)\,$\downarrow$ & Iters\,$\downarrow$ & Win\,$\uparrow$ & $p$ \\
    \midrule
    -- & Raw (no alignment)                    & 158.67 & 0.5703 & 0.1655 & -- & -- & -- & -- \\
    -- & After pre-alignment ($\mathcal{T}_0$) & 4.62   & 0.9439 & 0.6287 & -- & -- & -- & -- \\
    \midrule
    \multirow{2}{*}{500}  & Combined              & 4.11 & 0.9320 & 0.6679 & 0.94 & 59.8 & -- & -- \\
                          & Structure-constrained & \textbf{3.30} & \textbf{0.9435} & \textbf{0.7011} & \textbf{0.73} & \textbf{34.4} & 195/308 & $<0.001$ \\
    \midrule
    \multirow{2}{*}{1000} & Combined              & 2.73 & 0.9516 & 0.7426 & 1.46 & 42.2 & -- & -- \\
                          & Structure-constrained & \textbf{2.49} & \textbf{0.9537} & \textbf{0.7497} & \textbf{0.93} & \textbf{27.9} & 177/308 & $<0.001$ \\
    \bottomrule
  \end{tabular}
  }
\end{table}

\begin{figure}[t]
  \centering
  \includegraphics[width=0.6\textwidth]{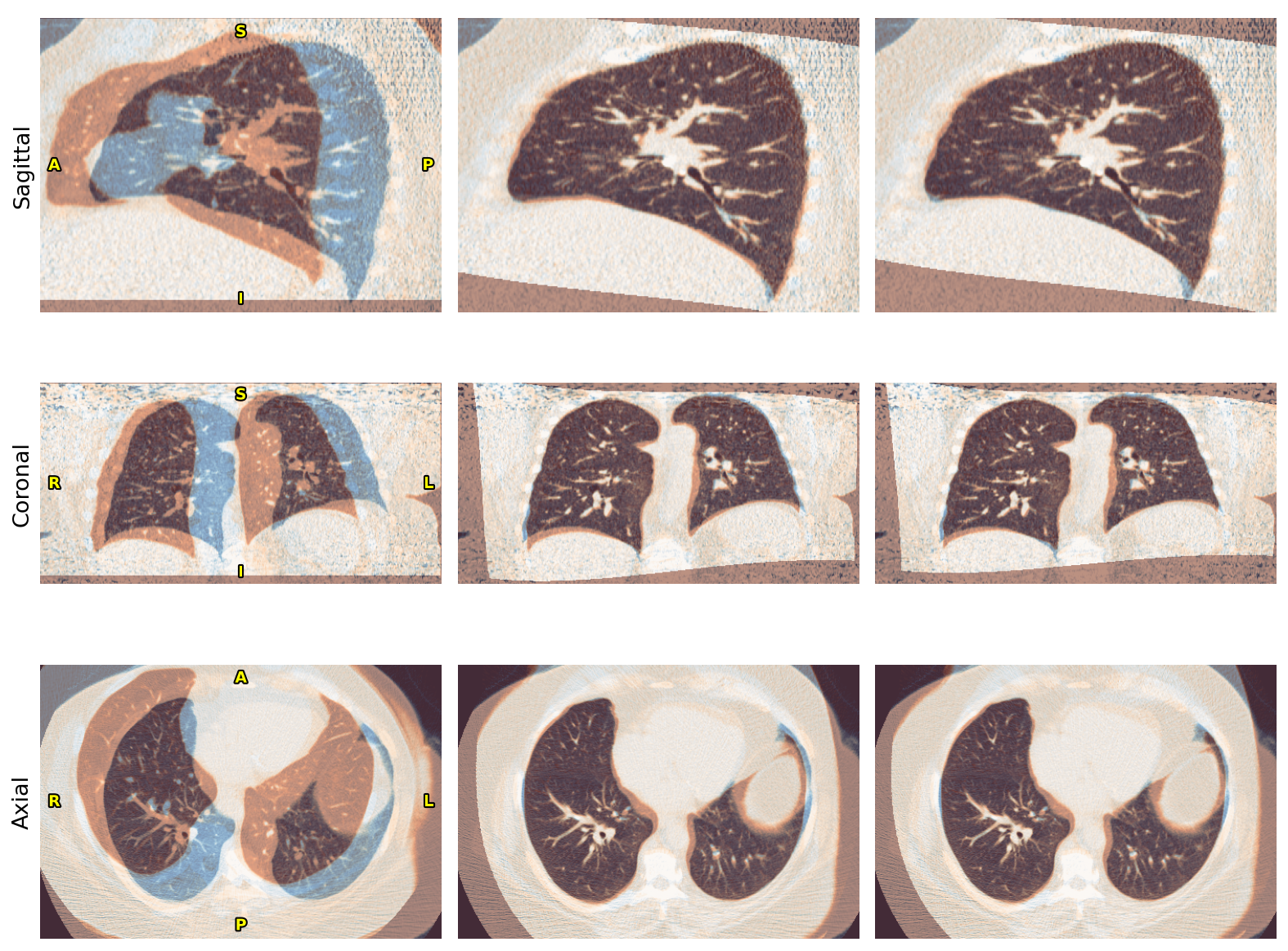}
  \caption{Qualitative comparison on a representative NLST pair at $N=1000$. Sagittal, coronal, and axial overlays of the warped fixed CT (orange) on the moving CT (blue). Gray indicates agreement, and residual orange/blue marks misalignment. In each row, the left, center, and right columns show the raw unregistered pair (no alignment), the combined-cloud baseline, and the proposed structure-constrained method, respectively.}
  \label{fig:qualitative}
\end{figure}

The structure-constrained correspondence improves every accuracy metric over the combined-cloud baseline at both budgets (Table~\ref{tab:results}), lowering nodule TRE by 0.81 mm at $N=500$ and 0.24 mm at $N=1000$ while raising lung and vessel Dice, and attaining the lower per-pair TRE on the majority of the 308 pairs ($p<0.001$). Both methods also improve on the pre-alignment reference (TRE 4.62 mm), except for lung Dice at $N=500$, where the correspondence is driven by interior vessel centerlines and constrains the lung surface only indirectly, leaving both methods below the reference (0.9439). The structure-constrained method stays nearest it (0.9435) and surpasses it at $N=1000$ (0.9537).



These results support the premise that cross-structure correspondence degrades the combined-cloud registration. Restricting transport to within-structure matches removes the artery–vein misalignment admitted by a purely geometric cost, and the joint warp step converts the per-structure correspondences into a single deformation without adding parameters. At the same total budget, the method is also faster. Since partitioning replaces the single $N\times N$ transport problem with $S$ problems of size $(N/S)\times(N/S)$ and converges in fewer iterations, the fit-time reduction arises from both fewer and cheaper iterations. 



Several limitations qualify these results. The constraint presumes a segmentation that separates arteries from veins, so labeling errors may propagate to the correspondence, and the nodule-TRE endpoint relies on a single landmark per pair. These motivate extending the partition to additional structures, such as the airways, and evaluating it on a denser landmark set.

\section{CONCLUSION}
We presented a structure-constrained correspondence for Iterative Optimal Transport (IOT) registration of pulmonary vessel point clouds, solving an unbalanced OT plan within each anatomical structure while fitting a single polynomial deformation jointly across structures with the same closed-form warp step on stacked inputs. The formulation adds no parameters over the base method. On intra-patient NLST chest CT pairs, it significantly reduces nodule target registration error and improves vessel and lung Dice, while reducing fit time and iteration count relative to the combined-cloud baseline at matched point budget. The modification is direct and low-cost, and applies when adjacent structures with distinct labels are registered.

\acknowledgments 
 
Research reported in this publication was supported by the NIH/National Cancer Institute Early Detection Research Network (U2CCA271898) and Liquid Biopsy Consortium (U01CA233370). The content is solely the responsibility of the authors and does not necessarily represent the official views of the National Institutes of Health.

\bibliography{report} 
\bibliographystyle{spiebib} 

\end{document}